\def\paperID{15809}
\def\confName{CVPR}
\def\confYear{2024}

\def\paperTitle{Physics-guided Deep Unfolding Network for Enhanced \\ Kronecker Compressive sensing}

\def\authorBlock{
    Gang Qu, Ping Wang, Siming Zheng, Xin Yuan \\
    Westlake University
}

\newif\ifreview 
\newif\ifarxiv \newcommand{\arxiv}{\arxivtrue}
\newif\ifcamera 
\newif\ifrebuttal 

\arxiv 

\pdfoutput=1
\documentclass[10pt,twocolumn,letterpaper]{article}
\ifreview \usepackage[review]{cvpr} \fi
\ifarxiv \usepackage[pagenumbers]{cvpr} \fi
\ifrebuttal \usepackage[rebuttal]{cvpr} \fi
\ifcamera \usepackage{cvpr} \fi

\usepackage{graphicx}	
\usepackage{amsmath}	
\usepackage{amssymb}	
\usepackage{booktabs}
\usepackage{times}
\usepackage{microtype}
\usepackage{epsfig}
\usepackage[table,xcdraw,dvipsnames]{xcolor}
\usepackage{caption}
\usepackage{float}
\usepackage{placeins}
\usepackage{color, colortbl}
\usepackage{stfloats}
\usepackage{enumitem}
\usepackage{tabularx}
\usepackage{xstring}
\usepackage{multirow}
\usepackage{xspace}
\usepackage{url}
\usepackage{subcaption}
\usepackage{xcolor}
\usepackage[hang,flushmargin]{footmisc}

\ifcamera \usepackage[accsupp]{axessibility} \fi

% ...

\ifarxiv  \fi

\newcommand{\R}[1]{{%
    \textbf{%
        \ifstrequal{#1}{1}{\textcolor{red}{R#1}}{%
        \ifstrequal{#1}{2}{\textcolor{blue}{R#1}}{%
        \ifstrequal{#1}{3}{\textcolor{magenta}{R#1}}{%
        \ifstrequal{#1}{4}{\textcolor{teal}{R#1}}{%
                           \textcolor{cyan}{R#1}%
        }}}}%
    }%
}}

\usepackage{xr-hyper}
\usepackage{cases}
\usepackage{amsmath}

\usepackage{amssymb}
\usepackage{url}            
\usepackage{booktabs}       
\usepackage{nicefrac}       
\usepackage{microtype}      
\usepackage{xcolor}         
\usepackage{graphicx}
\usepackage{caption}
\usepackage{tikz}
\usepackage{enumitem}
\usepackage{balance}
\usepackage{multirow}                
\usepackage{multicol}              
\usepackage{float}                   
\usepackage{makecell}                  
\usepackage{threeparttable}

\newcommand{\Amat}{{\boldsymbol A}}
\newcommand{\Bmat}{{\boldsymbol B}}

\newcommand{\Dmat}{{\boldsymbol D}}

\newcommand{\Fmat}[0]{{{\boldsymbol F}}}
\newcommand{\Gmat}[0]{{{\boldsymbol G}}}

\newcommand{\Kmat}[0]{{{\boldsymbol K}}}

\newcommand{\Omat}[0]{{{\boldsymbol O}}}
\newcommand{\Pmat}[0]{{{\boldsymbol P}}}
\newcommand{\Qmat}[0]{{{\boldsymbol Q}}}

\newcommand{\Tmat}[0]{{{\boldsymbol T}}}
\newcommand{\Umat}{{{\boldsymbol U}}}
\newcommand{\Vmat}[0]{{{\boldsymbol V}}}
\newcommand{\Wmat}[0]{{{\boldsymbol W}}}
\newcommand{\Xmat}{{\boldsymbol X}}
\newcommand{\Ymat}[0]{{{\boldsymbol Y}}}

\newcommand{\bv}{\boldsymbol{b}}

\newcommand{\gv}[0]{{\boldsymbol{g}}}

\newcommand{\xv}[0]{\boldsymbol{x}}
\newcommand{\yv}[0]{\boldsymbol{y}}

\newcommand{\Phimat}{\boldsymbol{\Phi}}
\newcommand{\Psimat}{\boldsymbol{\Psi}}

\newcommand{\ts}{^{\top}}

\newcommand{\vc}{{\rm vec}}
\newcommand{\argmin}{\arg\!\min}

\makeatletter
\newcommand*{\addFileDependency}[1]{
  \typeout{(#1)}
  \@addtofilelist{#1}
  \IfFileExists{#1}{}{\typeout{No file #1.}}
}

\makeatother
\newcommand*{\myexternaldocument}[1]{
    \externaldocument{#1}
    \addFileDependency{#1.tex}
    \addFileDependency{#1.aux}
}

\definecolor{cvprblue}{rgb}{0.21,0.49,0.74}
\usepackage[pagebackref,breaklinks,colorlinks,citecolor=cvprblue]{hyperref}
\usepackage[capitalize]{cleveref}
\crefname{section}{Sec.}{Secs.}
\crefname{table}{Table}{Tables}
\crefname{figure}{Fig.}{Figs.}

\ifarxiv \crefname{appendix}{App.}{Apps.}
\else \crefname{appendix}{Suppl.}{Suppls.} \fi

\unless\ifarxiv \myexternaldocument{_supplementary} \fi

\begin{document}
\title{\paperTitle}
\author{\authorBlock}
\maketitle

\begin{abstract}
Deep networks have achieved remarkable success in image compressed sensing (CS) task, namely reconstructing a high-fidelity image from its compressed measurement.
However, existing works are deficient in {\em incoherent compressed measurement} at sensing phase and {\em implicit measurement representations} at reconstruction phase, limiting the overall performance.
In this work, we answer two questions: $i)$ how to improve the measurement incoherence for decreasing the ill-posedness; $ii)$ how to learn informative representations from measurements.
To this end, we propose a novel asymmetric Kronecker CS (AKCS) model and theoretically present its better incoherence than previous Kronecker CS with minimal complexity increase.
Moreover, we reveal that the unfolding networks' superiority over non-unfolding ones result from sufficient gradient descents, called explicit measurement representations.
We propose a measurement-aware cross attention (MACA) mechanism to learn implicit measurement representations.
We integrate AKCS and MACA into widely-used unfolding architecture to get a measurement-enhanced unfolding network (MEUNet).
Extensive experiences demonstrate that our MEUNet achieves state-of-the-art performance in reconstruction accuracy and inference speed.

\end{abstract}
\section{Introduction}
\label{sec:intro}
Sensing is the way for human beings to detect and observe the world. How to effectively sense and recover the data approaching reality has always been a challenge for imaging. Compressive sensing (CS) is a cutting-edge paradigm which leverages the notion that natural signal can be sparsely represented in certain domains, allowing for efficient compression and subsequent reconstruction with minimal loss of information. The fundamental principle of CS lies in acquiring fewer samples than traditionally required by Nyquist-Shannon sampling theorem, while still capturing the essential information needed for reconstruction. This departures from conventional sampling paradigms has broad implications for various applications, including single-pixel cameras\cite{duarte2008single}, lensless imaging\cite{Yuan18OE, slope}, wireless remote monitoring\cite{zhang2012compressed}, snapshot compressive imaging\cite{9363502,wang2023efficientsci, qin2025detail}, and so on. 
\begin{figure}[ht]
  \centering
  \includegraphics[width=0.9\linewidth]{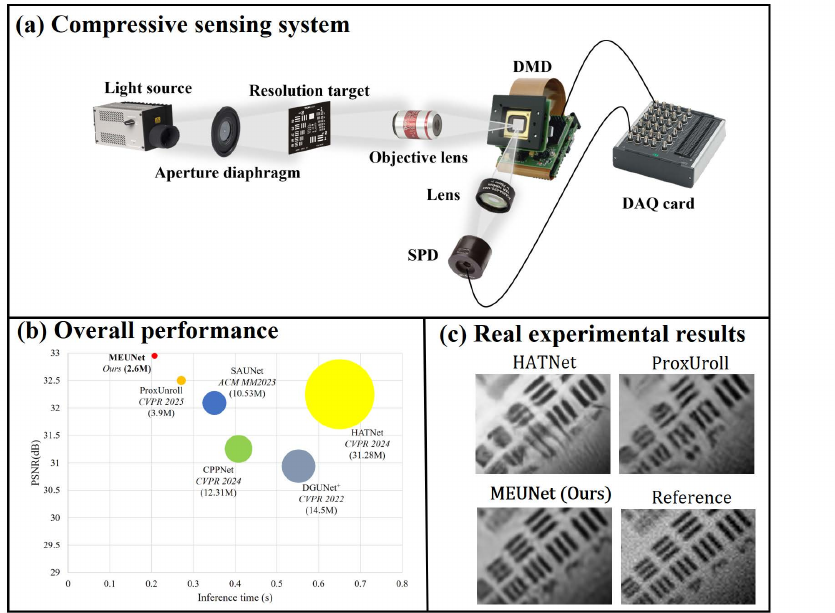}
  \caption{The proposed MEUNet achieves SOTA performance on both simulated and real data.}
\label{fig: Fig_1}
\end{figure}
CS theory has been widely studied in the past decades~\cite{candes2006robust,donoho2006compressed, duarte2011kronecker}, which indicates that the signal can be largely compressed and reconstructed under the Niquist sampling theorem. Concretely, the target 1D signal  $\xv \!\in\! \mathbb{R}^{N}$ can be linearly projected into a compressed measurement $\yv \!\in\! \mathbb{R}^{M}$ at a sub-Nyquist sampling ratio $\frac{M}{N}$ ($M \!\ll\! N $): $\yv = \Amat \xv + \epsilon$, where $\Amat \!\in\! \mathbb{R}^{M\times N}$ represents the sensing matrix, and $\epsilon$ denotes  noise.
The reconstruction of $\xv$ from the compressed measurement $\yv$ with $\Amat$ is an ill-posed problem,  which aims to solve the following optimization problem.
\begin{equation}
\label{eq:1}
\setlength{\abovedisplayskip}{0.25cm}
\setlength{\belowdisplayskip}{0.25cm}
\textstyle \hat{\xv} = \mathop{\arg\min}\limits_{\xv}\frac{1}{2}\Vert \yv - \Amat\xv \Vert^2_2+\lambda \gv(\xv), 
\end{equation}
where $\frac{1}{2}\Vert \yv-\Amat\xv \Vert^2_2$ is the data fidelity term, and $\lambda\gv(\xv)$ is the regularization term.
To solve the above optimization problem, either conventional iterative algorithms~\cite{potter2010sparsity, sparsity-dl, GAPTV, rank}, or deep learning-based methods~\cite{kulkarni2016reconnet,lyu2017deep,shi2019scalable,yao2019dr2} can be considered as solvers. Thus, for CS tasks, two important factors heavily affect the performance of reconstruction. In the sensing phase, how to capture the target more efficiently and effectively. In the reconstruction phase, how to design an advanced algorithms to further explore the correlations among measurement and sensing matrix. 
Deep unfolding networks (DUNs) have been proposed for CS reconstruction in the past few years, which not only enjoy both interpretability and excellent performance in quality and flexibility, but jointly optimized the sensing matrix for detection. However, the introduction of gradient projection in DUN inevitably leads to the decline in efficiency, especially using the large sensing matrix whose size increases exponentially with image scale. Block CS (BCS) is one solution to avoid the use of the large sensing matrix, which separate the whole image into non-overlapping patches, and then implement CS patch by patch, thus largely improving the efficiency. However, problems are also introduced when using BCS: $i)$ The global interactions among whole image are lost, thus limiting the performance in reconstruction. $ii)$ Block CS breaks the laws of imaging in real optical system, restricting the real-world application for image CS. 

Bearing these concerns in mind, Kronecker CS (KCS) has been adopted in DUN methods in the most recent works~\cite{wang2023saunet, qu2024dual}, which aims to solve the problems in block CS. Different from block CS, KCS is a full-size image modulation method with no need to separate the image into patches, but uses the Kronecker product of two sub-matrices (image scale) to replace the conventional large modulation matrix. The mathematical model can be formulated as following: 
\begin{equation}
\label{eq:kcs}
\Ymat = \Phimat \Xmat \Psimat^\top,
\end{equation}
where $\Ymat \in \mathbb{R}^{m\times n}$ is the 2D compressed measurement, $\Xmat \in \mathbb{R}^{H\times W}$ is the 2D image, $\Phimat \in \mathbb{R}^{m\times H}$ and $\Psimat \in \mathbb{R}^{n\times W}$ are two independent sub-matrices, compression ratio is defined as $\frac{mn}{HW}$.
Based on the properties of the Kronecker product, the Eq.~\eqref{eq:kcs} is equivalent to the following vectorized form:
$\vc \left(\Ymat\right) = \left( \Psimat \otimes \Phimat \right) \vc \left(\Xmat\right)$,
where vec(·) denotes the vectorization operation, and $\otimes$ is the Kronecker product. This kind of transformation indicates that the KCS can replace traditional CS in most situations when the large sensing matrix is not necessary to be pre-determined, so we can use two sub-matrices to replace the large one. Thus, KCS totally solve the problems the block CS introduced, which conducts CS in full 2D image level, and is also applicable in real-world CS systems, e.g., SPI system~\cite{qu2024dual}. KCS also provides high efficiency because the forward and inverse gradient projection process in DUN are also implemented at image level. However, the drawbacks of KCS have not been discussed yet in previous works.
The rigid structure of the Kronecker product imposes a strong coupling relation between sub-matrices, leading to the large sensing matrix with high mutual coherence and suboptimal restricted isometry property (RIP) constants. This structural prior, while computationally efficient, fundamentally limits the expressiveness of the learned sensing operator in DUN and, consequently, limited reconstruction quality. Thus, this is the measurement barrier at the sensing end.
To address this limitation, we propose asymmetric Kronecker CS (AKCS), a novel sensing paradigm that retains the computational efficiency of KCS while dramatically enhancing its capability of expressiveness in DUN. AKCS conditions the column-wise sensing matrix on the choice of the row-wise sensing basis, effectively assigning a unique, learnable column operator for each row measurement, which breaks the restrictive symmetry of the Kronecker product.
In the reconstruction end, we further break the measurement barrier by exploring the implicit prior of measurement, rather than simply explicit representations as a data fidelity term in DUN. 
Overall, our contributions are summarized as follows:
\begin{itemize}
\item 
We propose AKCS model, which breaks the limitations of standard Kronecker CS paradigm by introducing asymmetric row-adaptive sub-matrices. We theoretically analyze the coherence of the sensing matrix generated by the standard KCS and AKCS model, and present the demonstration of its superiority in the capability of expressiveness in DUN.
\item We demonstrate the compressive measurement is an important prior for CS reconstruction in DUN model. A measurement-aware cross-attention module is specifically designed to capture and fuse the global feature prior embedded in measurement, leading to an efficient and powerful DUN model for CS reconstruction.  
\item The combination of AKCS model and measurement-aware cross-attention module in DUN leads to MEUNet, which can be applied for fast and high-accuracy CS reconstruction tasks. Further experiments on simulated dataset and real data demonstrate the performance of the proposed method, which also indicates its potential to be applied in various real-world CS systems for efficient and high-quality reconstruction. 
\end{itemize}

\section{Related Work}
\label{sec:related}
\subsection{Image CS Reconstruction}
CS reconstruction methods can be classified into two categories: model-based methods~\cite{figueiredo2007gradient,4587391,he2009exploiting,blumensath2009iterative,beck2009fast,kim2010compressed,yang2011alternating,dong2014compressive,zhang2014group,Metzler2016FromDT} and learning-based methods~\cite{kulkarni2016reconnet,metzler2017learned,zhang2018ista,yang2018admm,shi2019image,shi2019scalable,yao2019dr2,zhang2020optimization,zhang2020amp,shen2022transcs,song2021memory,mou2022deep,song2023optimization}. The conventional model-based methods mainly rely on some hand-crafted priors to recover the original image from its sub-sampling measurement in an iterative manner, which 
enjoy high generalization and robustness, but meanwhile suffer from high computational cost and limited reconstruction quality. 
Earlier deep learning-bsed methods~\cite{kulkarni2016reconnet,metzler2017learned} treat DNN as a black box and  directly build a mapping from compressed measurement to the image. Recently, DUNs are proposed to incorporate DNN with conventional model-based methods, and train the unfolding network with multiple stages in an end-to-end manner, which enjoys good interpretability and has become the mainstream for CS reconstruction. Different optimization methods lead to different optimization-inspired DUNs, e.g., proximal gradient descent (PGD) algorithms~\cite{chen2022content, song2021memory}, AMP~\cite{zhu2020deformable}, ADMM~\cite{Wang_2025_CVPR} and so on. 
Although DUN presents superiority compared to the end-to-end design, the introduction of gradient projection inevitably increases the computational burden, especially for large-scale image with corresponding large sensing matrix. KCS is a possible solution for this issue and has been introduced into DUN, which shows better performance due to the retaining of global information. KCS-based DUN leads to SOTA performance for CS reconstruction and meanwhile maintains high efficiency, and most recent works have extend KCS-based model into real-world CS systems~\cite{qu2024dual}. However, the highly structured distribution of sensing matrix caused by the KCS actually restrict its performance in DUN, which has not been discussed yet in previous works.

\section{Method}
\label{sec:method}

\begin{figure*}[t]
  \centering
  \includegraphics[width=0.8\linewidth]{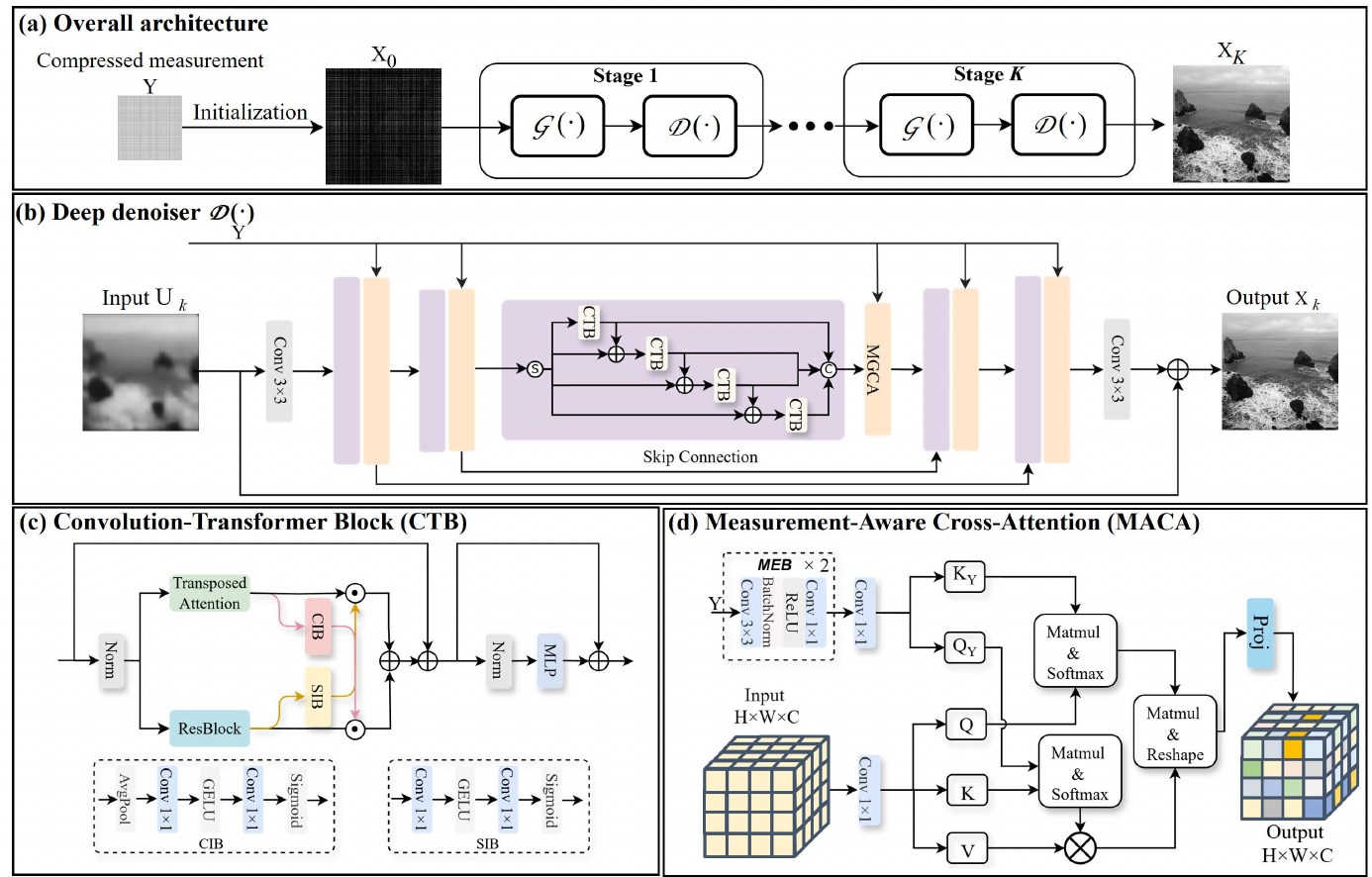}
  \caption{The pipeline of the proposed method. (a) The overall architecture of the DUN model, containing $K$ stages. Each stage includes proximal gradient descent projection of Eq.~\eqref{eq: v} and trainable deep deniser of Eq.~\eqref{eq: akxx}. (b) The design of deep denoiser in each stage, and all stages share the same parameters of deep denoiser. (c) The convolution-Transformer block in deep denoiser. (d) The design of measurement-aware cross-attention block.}
\label{fig: overall}
\end{figure*}

\subsection{Asymmetric Kronecker CS}
Stand KCS introduces a measurement barrier at the sensing end due to the strong correlations induced by the Kronecker product structure.
While the proposed AKCS can solve the coupling between row and column operations in KCS by imposing asymmetric row-adaptive design of sub-matrices. 
AKCS ensures that no two rows share a common structural component, thus the number of correlated pairs is drastically reduced. 
The forward model of AKCS can thus be expressed as:
\begin{equation}
\label{eq:ara}
\Ymat = {\cal C}_{i=1:m}[\Phimat_i \Xmat \Psimat_i^\top].
\end{equation}
where ${\cal C}$ denotes {\em concatenation operation} along row dimension, $\Phimat_i \in \mathbb{R}^{1\times H}$ and $\Psimat_i \in \mathbb{R}^{n \times W}$ are row-wise matched matrices. 
In this way, the coupling relation of $\Phimat$ and $\Psimat$ in standard KCS is broken, which means that the coherence of the generated large sensing matrix would largely drop and further provides more freedom in DUN for optimization. In the following part, we theoretically analyze the coherence to demonstrate the superiority of AKCS.

\noindent{\bf Coherence Analysis of KCS and AKCS.}
In CS, coherence is an important factor to measure the expressive power of the sensing matrix, which is defined as the maximum absolute inner product of the normalized column vectors of the matrix, i.e., the absolute value of the maximum off-diagonal element of the Gram matrix:
\begin{equation}
\label{eq:coherence}
\mu_A = {\rm max}_{i\neq j}|<a_i, a_j>|,
\end{equation}
where $a_i$ denotes the $i$-th column of $A$. Theoretically, the sensing matrix with lower coherence guarantees a better reconstruction quality with less number of measurement. Then, we will analyze the coherence of generated sensing matrices $\Amat_K$ using KCS and $\Amat_{AK}$ using AKCS model.

(i) The coherence of $\Amat_K$.

The Gram matrix of Kronecker product has a closed-form function. Considering the Kronecker product of $\Amat_K = \Omat \otimes \Pmat$, where $\Omat \in \mathbb{R}^{m\times H}$, and $\Pmat \in \mathbb{R}^{n\times W}$. Its Gram matrix can be expressed as:
\begin{align}
\label{eq:GK}
\Gmat_K = \textstyle\Amat_K{\ts} \Amat_K = (\Omat \otimes \Pmat){\ts} (\Omat \otimes \Pmat) \\ \notag
=  \textstyle(\Omat{\ts} \Omat) \otimes (\Pmat{\ts} \Pmat) = \Dmat \otimes \Bmat,
\end{align}
The elements of $\Gmat_K$ can be represented as $\Gmat_K[(i,k), (j,l)] = \Bmat_{i,j}\Dmat_{k,l}$, where the indexes are the positions of the corresponding vectorized form. Assume the columns of $\Pmat$ and $\Omat$ have been normalized, then $\Bmat_{i,i}=1$ and $\Dmat_{i,i}=1$. The coherence of $\Amat_K$ is:
\begin{align}
\label{eq:uk}
\mu_K = & \textstyle\underset{(i,k)\neq (j,l)}{\rm max}|\Bmat_{i,j}, \Dmat_{k,l}| \\ \notag
 =  & \textstyle{\rm max}(\underset{i\neq j}{\rm max}|\Bmat_{i,j}|,  \underset{k\neq l}{\rm max}|\Dmat_{k,l}|,    \underset{i\neq j, k\neq l}{\rm max}|\Bmat_{i,j}\Dmat_{k,l}|). 
\end{align}
Because $|\Bmat_{i,j}| \leq \mu_{\Pmat} = \underset{i\neq j}{\rm max}|\Bmat_{i,j}|$, and $|\Dmat_{k,l}| \leq \mu_{\Omat} = \underset{k\neq l}{\rm max}|\Dmat_{k,l}|$, and  $|\Bmat_{i,j}\Dmat_{k,l}| \leq \mu_{\Pmat}\mu_{\Omat}$, while $\mu_{\Pmat}\mu_{\Omat} \leq {\rm max}(\mu_{\Pmat}, \mu_{\Omat})$, thus
\begin{equation}
\label{eq:uk2}
\mu_K = {\rm max}(\mu_{\Pmat}, \mu_{\Omat}),
\end{equation}
where $\mu_{\Pmat}$ and $\mu_{\Omat}$ represent the coherence of $\Pmat$ and $\Omat$, respectively. 

Assume the elements in $\Omat$ and $\Pmat$ are independent identically distributed (i.i.d.), and follow Gaussian distribution $\sim \mathcal{N}(0, 1)$, the coherence is  approximately presented as:
\begin{equation}
\label{eq:coh}
\textstyle
\mu_{\Omat} = \sqrt{\frac{2\log H}{n}},\quad \mu_{\Pmat} = \sqrt{\frac{2\log W}{m}} , 
\end{equation}
where $H$ and $W$ denote the number of columns of $\Omat$ and $\Pmat$, $m$ and $n$ are the number of rows.

(ii) The coherence of $\Amat_{AK}$.

Unlike KCS, $\Amat_{AK}$ is generated by the asymmetric row-adaptive sub-matrices, whose columns can be represented as:
\begin{equation}
\label{eq:arwcs}
{c_{i,j} = [a_{1,j} \bv_{1,i}, a_{2,j} \bv_{2,i}, ...]\ts},
\end{equation}
where $a_{m,j}$ denotes the elements in $\Phimat$, $\bv_{m,i}$ is the $i$-th column of $\Psimat$. Thus, the Gram matrix can be expressed as:
\begin{equation}
\label{eq:gm}
\scalebox{0.85}{$
\Gmat_{AK}[(i,j), (k,l)] = <c_{i,j}, c_{k,l}> = \sum_{r = 1}^{m}{a_{r,j}a_{r,l}<\bv_{r,i}, \bv_{r,k}>}$}
\end{equation}
The coherence of $\Amat_{AK}$ can be obtained:
\begin{equation}
\label{eq:cak}
\scalebox{0.95}{$
\textstyle
\mu_{AK} = \underset{(i,j)\neq (k,l)}{\rm max} \frac{|\sum_{r = 1}^{m}{a_{r,j}a_{r,l}<\bv_{r,i}, \bv_{r,k}>|}}{\sqrt{\sum_{m}a^2_{m,j}||\bv_{m,j}||_2^2} \sqrt{\sum_{m}a^2_{m,l}||\bv_{m,k}||_2^2}}.
$}
\end{equation}
The situation of $\mu_{AK}$ is more complicated than $\mu_{A}$, we directly present conclusion, summarized as {\bf Theorem 1}:

\noindent{\bf Theorem 1}. Assume a sensing matrix 
\begin{equation}
\label{eq:sm}
A_{AK}=[a_1 \otimes B_1,...,a_m \otimes B_m]\ts,
\end{equation}
where $a_i \in \mathbb{R}^{1\times H}$ follows i.i.d., $\sim \mathcal{N}(0, 1)$, $B_i \in \mathbb{R}^{n\times W}$ follows i.i.d., $\sim \mathcal{N}(0, 1)$, there exists an absolute constant $C_o > 0$ ($C_o\approx 1$ for Gaussian distribution), such that with high probability:
\begin{equation}
\label{eq:ukpq}
\textstyle
\mu_{AK} \leq C_o \sqrt{\frac{2\log(HW)}{mn}}.
\end{equation}
The theoretical demonstration of \textbf{Theorem 1} is presented in Supplementary material $\mathcal S1$.
Thus, we get the upper bounds of $\mu_K$ and $\mu_{AK}$, and obviously, $\mu_{AK} <\mu_{K}$. This means the coherence of sensing matrix generated by AKCS is much less than that generated by KCS, which indicates that the AKCS model is potential to promote the CS reconstruction in DUN.

\noindent{\bf AKCS model in DUN.}
Bearing in mind the theoretical analysis above, we adopt AKCS in DUN model, and consider the optimization problem of Eq.~(\ref{eq:ara}), which can be rewritten as:
\begin{equation}
\label{eq: dec}
\scalebox{0.95}{$
 {\hat \Xmat} =\textstyle{\argmin}_{\Xmat}~\frac{1}{2} \left \| {\Ymat} - {\cal C}_{i=1:m}[\Phimat_i \Xmat \Psimat_i^\top] \right \|_{F}^{2}+ \lambda \gv({\Xmat}),
 $}
\end{equation}
where ${\left\|  \cdot  \right\|_{F}}$ denotes the Frobenius norm.
Further developing ISTA algorithms to solve~\eqref{eq: dec}, then we have:

\begin{equation}
\begin{aligned}
{\label{eq: v}
{\Umat_{k}  =  \textstyle \Xmat_{k-1} + \rho  {\sum}_{i=1:m} [{\Phimat_i^\top(\Ymat_i - \Phimat_i \Xmat_{k-1} \Psimat_i^\top)\Psimat_i}],}}\\
\end{aligned}
\end{equation}
\begin{equation}
\begin{aligned}
{\label{eq: xx}
{{\Xmat}_{k} = \textstyle {\argmin}_{\Xmat}~\frac{1}{2{\sigma^2}} || \Umat_{k} - \Xmat  ||_{F}^{2} + \gv (\Xmat),}}
\end{aligned}
\end{equation}
where $\Ymat_i$ denotes the $i$-th row of 2D measurement $\Ymat$. Eq.~\eqref{eq: v} denotes the gradient descent from the current stage input $\Xmat_{k-1}$ with a step size $\rho$, $\Umat_{k}$ is an auxiliary variable. Eq.~\eqref{eq: xx} denotes proximal mapping process, which can be seen as a denoising problem under the noise level $\sigma$. In DUN, this denoising problem is conducted by a trainable deep neural network, which has shown its superiority compared to those hand-designed denoisers. When integrating the ISTA algorithm with a specific denoising network, a AKCS-DUN model can be established:
\begin{equation}
\begin{aligned}
{\label{eq: akv}
{\Umat_{k}  =  \Xmat_{k-1} + \rho_{k-1}  {\sum}_{i=1:m} [ {\Phimat_i^\top(\Ymat_i - \Phimat_i \Xmat_{k-1} \Psimat_i^\top)\Psimat_i}],}}\\
\end{aligned}
\end{equation}
\begin{equation}
\begin{aligned}
{\label{eq: akxx}
{{\Xmat}_{k} =  {\mathcal D}_{(k,\theta)}(\Umat_{k})},}
\end{aligned}
\end{equation}

where $\rho_{k-1}$ is a learnable parameter controls the step size in each stage, and ${\mathcal D}_{(k,\theta)}$ represents the learnable deep denoiser.

\subsection{Design of Deep Denoiser}
In the deep denoising phase, we aim to learn informative implicit representations from previous output $\Umat_k$ and the measurement $\Ymat$.
Next, 
the overall architecture of the proposed deep denoiser is first given and then two core modules, hybrid Convolution-Transformer block (CTB) and measurement-aware cross-attention (MACA) block are introduced in detail.

\noindent{\bf Overall architecture.} 
As shown in Fig.~\ref{fig: overall} (b), our deep denoiser ${\mathcal D}$ is a symmetric UNet architecture for multi-scale representation learning, composed of two encoder layer, a bottleneck layer, and two decoder layer. 
The input of deep denoiser is the output $\Umat_{k}$ of the gradient descent, which then passes through encoders to generate the deep features. In each layer, the features $\Fmat_{k-1}$ of current layer are concatenated with features from previous stage ${k-1}$ and fused by channel concatenation and $1\times1$ convolution. To  reduce the computational complexity, we separate the input features into four groups along channel dimension and adopt the residual connection for interactions among them. The encoder in each layer mainly contains convolution-Transformer block (CTB), measurement-aware cross-attention (MACA) block, and down-sampling layer. The goal of encoders is to progressively reduce the spatial resolution by half and double the channel dimensions, yielding the multi-scale features transferred to the decoder by skip connections. In the decoder branch, the same design of encoder is adopted, with simple $1\times1$ convolution and pixel shuffle operations for inner-stage feature fusion and up-sampling. The output of each decoder $\Fmat_k$ is the feature map of the current stage and are then transferred to the next stage. Finally, the feature map from the last decoder is converted to the output $\Xmat_k \!\in\! \mathbb{R}^{N_H \times N_W}$ of current stage with a $3\times3$ convolution.

\noindent{\bf Convolution-Transformer Block.} 
The mixture modules of convolution and Transformer is a common design to maintain the local and non-local modeling capability of deep neural network. 
Considering the efficiency and the following design of cross-attention in spatial dimension, we adopt the parallel blocks of convolution and channel-wise self-attention with cross interactions here.
As depicted in Fig.~\ref{fig: overall}(c), the CTB mainly contains spatial convolution block (SCB), residual-squeeze channel attention (RSCA) block, interaction blocks. 
Considering the efficiency, SCB contains only two $3\times3$ convolution with LeakyReLU inside, which captures the local interactions effectively, and the output of SCB $\Fmat_{s} \!\in\! \mathbb{R}^{N_H\times N_W\times {N_C}}$ is just the same size with input $\Fmat_{in}$. 

As for RSCA, considering the same input $\Fmat_{in} \!\in\! \mathbb{R}^{N_H \times N_W\times N_C}$, it is first reshaped to $\Fmat_{cr} \!\in\! \mathbb{R}^{N_HN_W\times N_C}$. Then the Query ($\Qmat$), Key ($\Kmat$), and Value ($\Vmat$) can be obtained by linear transform:
\begin{equation}
\begin{aligned}
\label{eq: CSFF1}
\Qmat = \Fmat_{cr} \Wmat^{q}, \Kmat = \Fmat_{cr} \Wmat^{k}, \Vmat = \Fmat_{cr} \Wmat^{v},
\end{aligned}
\end{equation}
where $\Wmat^{(\cdot)} \!\in\! \mathbb{R}^{ N_C \times N_C }$ denotes the learnable linear projection. Then the multi-head attention mechanism is now:
\begin{equation}
\begin{aligned}
\label{eq: mha}
\Fmat_{i} = \Amat_i * \Vmat_i = \textstyle {\rm softmax}(\frac{\Qmat_i\Kmat^\Tmat_i}{\sqrt{d}})*\Vmat_i,
\end{aligned}
\end{equation}
$i=1, \dots,N$ is the number of heads, ${d}$ is a learnable scaling parameter that controls the magnitude of the product. 

RSCA module further introduces a residual high-frequency enhancement branch. The core intuition is using 'squeeze' operation to aggregate the feature map along channel dimension, which means the global channel average and represents the low-frequency information (dominant component of each feature channel). The difference between the original feature and its low-frequency component constitutes the high-frequency residual. 
By explicitly isolating and then adaptively amplifying this residual, our module forces the network to pay specific attention to and preserve these high-frequency details, which are often lost in standard attention mechanisms:
\begin{equation}
\begin{aligned}
\label{eq: CSFF2}
\Fmat_{RSi} =\Fmat_{i} + \gamma (\Fmat_{i} - \phi(S(\Fmat_{i}))),
\end{aligned}
\end{equation}
where $\phi$ represents GELU activation function, $S$ denotes `squeeze' operation to squeeze channel dimension from $C$ to $1$, $\gamma$ is a learnable scaling factor. After a linear transform, we obtain the final output of RSCA block $\Fmat_{co} \!\in\! \mathbb{R}^{N_H\times N_W\times N_C}$. The output of both SCB and CSAB are concatenated for feature fusion along with a residual input, 
\begin{equation}
\begin{aligned}
\label{eq: out}
\hat{\Fmat_{o}} = {\rm Concat}(\Fmat_{s}, \Fmat_{c}) + \Fmat_{in},
\end{aligned}
\end{equation}
Following the regular design of FFNs in Transformer, With input $\hat{\Fmat_{o}}\!\in\! \mathbb{R}^{N_H \times N_W\times N_C}$, the output of FFN is:
\begin{equation}
\begin{aligned}
\label{eq: ffn}
{\Fmat_{o}} = \hat{\Fmat_{o}} + W_{(1\times1)}(W_d(\phi(W_{(3\times3)}(\hat{\Fmat_{o}})))),
\end{aligned}
\end{equation}
where $W_{(1\times1)}$ and $W_{(3\times3)}$ represent $1\times1$ convolution and $3\times3$ convolution, $W_d$ denotes the depth-wise convolution respectively.

\noindent{\bf Measurement-aware Cross-Attention Block.} 
The compressive measurement $\Ymat$ in image CS task is actually a global feature extracted from the ground truth with the sensing matrix, and the masks actually serve as global convolution kernels. Bearing this perspective in mind, we specifically design a measurement-aware cross-attention (MACA) block to further explore the implicit prior embedded in measurement and adopt it to guide the reconstruction in DUN.
In MACA, we treat the high-level information encoded from the measurements $\Ymat$ as a set of Queries and keys. They are then used to establish an information bridge from the measurement domain to the image feature domain. To strike a balance between prohibitive computational costs and the need for precise information flow, MACA employs a highly efficient `Aggregate-in-Low, Propagate-in-High' strategy. This ensures that global guidance is both potent and computationally tractable, even when operating on high-resolution feature maps.
The proposed MACA module is illustrated in Fig.~\ref{fig: overall} (d). The  computational flow is organized into two stages:(1)Forward Attention for Information Aggregation, and (2) Backward Attention for Information Propagation.
Let the input feature map be $\Fmat_{in} \!\in\! \mathbb{R}^{N_H\times N_W\times N_C}$, and the compressive measurement be $\Ymat \!\in\! \mathbb{R}^{N_h\times N_w}$.
The primary goal of stage (1) is to efficiently summarize the rich spatial information in $\Fmat_{in}$ under the guidance of the measurement-derived queries.
The measurement $\Ymat$ is first processed by a lightweight CNN, the Measurement Encoder Block (MEB), to produce a sequence of query vectors $\Qmat_Y \!\in\! \mathbb{R}^{N_Y \times N_d}$ and key vectors $\Kmat_Y \!\in\! \mathbb{R}^{N_Y \times N_d}$. This step transforms the raw, low-level measurements into a set of high-level, semantic ``questions" and `keys' about the image content.
To circumvent the quadratic complexity of attention on high-resolution feature maps, we then down-sample the input feature map $\Fmat_{in} \!\in\! \mathbb{R}^{N_H\times N_W\times N_C}$ by a factor of $D$  using average pooling, resulting in $\Fmat_{MD} \!\in\! \mathbb{R}^{\frac{N_H}{D} \times \frac{N_W}{D} \times N_d}$. This drastically reduces the sequence length of the keys and values from $N_HN_W$ to $\frac{N_HN_W}{D^2}$, directly tackling the primary computational and memory bottleneck.
Then we compute the low-resolution attention scores between the measurement queries $\Qmat_Y$ and the keys $\Kmat_D$ derived from the down-sampled features, which is used to weight the values $\Vmat_D$.
\begin{equation}
\begin{aligned}
\label{eq: atteny}
\Vmat_{YD} = \textstyle {\rm softmax}(\frac{\Qmat_Y\Kmat^\Tmat_D}{\sqrt{d_k}}) * \Vmat_D.
\end{aligned}
\end{equation}
Here, $\Kmat_D$, $\Vmat_D$ are linearly projected from $\Fmat_{MD}$. The output, $\Vmat_{YD} \!\in\! \mathbb{R}^{N_Y \times N_d}$, represents a compact summary of the image features, where each element is an ``answer" to the corresponding query from $\Ymat$.

\begin{table*} [ht]
\centering
\caption{Average PSNR/SSIM of different methods on Set11 datasets with different SRs. The best and second best results are highlighted in {\bf bold} and \underline{underlined}, respectively.}
\label{tab: result}
\renewcommand\tabcolsep{6pt}
\scalebox{0.92}{
\begin{tabular}{c|c|cccc}
\toprule[1pt]
\multicolumn{1}{c|}{\multirow{2}{*}{Dataset}}  &
\multicolumn{1}{c|}{\multirow{2}{*}{Method}}   & 
\multicolumn{4}{c}{Sampling Ratio (SR)} 
\\ \cline{3-6}  
\multicolumn{1}{c|}{} &  
\multicolumn{1}{c|}{}  &

\multicolumn{1}{c}{4\%}  & 
\multicolumn{1}{c}{10\%}  & 
\multicolumn{1}{c}{25\%}  & 
\multicolumn{1}{c}{50\%} \\
\toprule[1pt]
\multicolumn{1}{c|}{}  &
\multicolumn{1}{l|}{ReconNet~(CVPR 2016)}   & 

\multicolumn{1}{c}{20.93/0.5897} & 
\multicolumn{1}{c}{24.38/0.7301} & 
\multicolumn{1}{c}{28.44/0.8531} & 
\multicolumn{1}{c}{32.25/0.9177} \\
\multicolumn{1}{c|}{}  &
\multicolumn{1}{l|}{ISTA-Net$^+$~(CVPR 2018) }      & 

\multicolumn{1}{c}{21.32/0.6037} & 
\multicolumn{1}{c}{26.64/0.8087} & 
\multicolumn{1}{c}{32.59/0.9254} & 
\multicolumn{1}{c}{38.11/0.9707} \\
\multicolumn{1}{c|}{}  &
\multicolumn{1}{l|}{CSNet$^+$~(TIP 2019)}      & 

\multicolumn{1}{c}{24.83/0.7480} & 
\multicolumn{1}{c}{28.34/0.8580} & 
\multicolumn{1}{c}{33.34/0.9387} & 
\multicolumn{1}{c}{38.47/0.9796} \\
\multicolumn{1}{c|}{}  &
\multicolumn{1}{l|}{SCSNet~(CVPR 2019)}      & 

\multicolumn{1}{c}{24.29/0.7589} &
\multicolumn{1}{c}{28.52/0.8616} & 
\multicolumn{1}{c}{33.43/0.9373} &
\multicolumn{1}{c}{39.01/0.9769} \\
\multicolumn{1}{c|}{}  &
\multicolumn{1}{l|}{OPINE-Net$^+$~(JSTSP 2020))} &

\multicolumn{1}{c}{25.69/0.7920} &
\multicolumn{1}{c}{29.81/0.8884} & 
\multicolumn{1}{c}{34.86/0.9509} & 
\multicolumn{1}{c}{40.17/0.9797} \\
\multicolumn{1}{c|}{Set11}  &
\multicolumn{1}{l|}{AMP-Net~(TIP 2020)} &

\multicolumn{1}{c}{25.27/0.7821} &
\multicolumn{1}{c}{29.43/0.8880} & 
\multicolumn{1}{c}{34.71/0.9532} & 
\multicolumn{1}{c}{40.66/0.9827} \\
\multicolumn{1}{c|}{}  &
\multicolumn{1}{l|}{TransCS~(TIP 2022)}   & 

\multicolumn{1}{c}{25.41/0.7883	} & 
\multicolumn{1}{c}{29.54/0.8877} & 
\multicolumn{1}{c}{35.06/0.9548} & 
\multicolumn{1}{c}{40.49/0.9815} \\
\multicolumn{1}{c|}{}  &
\multicolumn{1}{l|}{MADUN~(ACM MM 2021))} &

\multicolumn{1}{c}{25.71/0.8042} &
\multicolumn{1}{c}{30.20/0.9016} & 
\multicolumn{1}{c}{35.76/0.9601} & 
\multicolumn{1}{c}{41.00/{0.9837}} \\
\multicolumn{1}{c|}{}  &
\multicolumn{1}{l|}{{DGUNet$^+$}~(CVPR 2022)} &
\multicolumn{1}{c}{{26.82/0.8230}} &
\multicolumn{1}{c}{{30.93/0.9088}} & 
\multicolumn{1}{c}{{36.18/0.9616}} & 
\multicolumn{1}{c}{41.24/{0.9837}} \\
\multicolumn{1}{c|}{}  &
\multicolumn{1}{l|}{OCTUF$^+$~(CVPR 2023)} &
\multicolumn{1}{c}{26.54/0.8150} &
\multicolumn{1}{c}{30.73/0.9036} & 
\multicolumn{1}{c}{36.10/0.9607} & 
\multicolumn{1}{c}{41.35/\underline{0.9838}} \\

\multicolumn{1}{c|}{}  &
\multicolumn{1}{l|}{SAUNet~(ACM MM 2023))} &
\multicolumn{1}{c}{{27.80}/{0.8353}} &
\multicolumn{1}{c}{{32.15}/{0.9147}} & 
\multicolumn{1}{c}{{37.11}/{0.9628}} & 
\multicolumn{1}{c}{{41.91}/\underline{0.9838}} \\

\multicolumn{1}{c|}{}  &
\multicolumn{1}{l|}{{CPPNet~(CVPR 2024))}}      & 
\multicolumn{1}{c}{{27.23}/{0.8337}} & 
\multicolumn{1}{c}{{31.27}/{0.9135}} & 
\multicolumn{1}{c}{{36.35}/{0.9631}} &
\multicolumn{1}{c}{-}  \\

\multicolumn{1}{c|}{}  &
\multicolumn{1}{l|}{{HATNet~(CVPR 2024))}}      & 
\multicolumn{1}{c}{{27.98}/{0.8382}} & 
\multicolumn{1}{c}{{32.26}/{0.9182}} & 
\multicolumn{1}{c}{{37.24}/{0.9634}} &
\multicolumn{1}{c}{{\underline{42.05}}/\underline{0.9838}} \\

\multicolumn{1}{c|}{}  &
\multicolumn{1}{l|}{{ProxUnroll~(CVPR 2025))}}      & 
\multicolumn{1}{c}{\underline{28.30}/\underline{0.8452}} & 
\multicolumn{1}{c}{\underline{32.55}/\underline{0.9226}} & 
\multicolumn{1}{c}{\underline{37.35}/\underline{0.9639}} &
\multicolumn{1}{c}{{{41.97}}/\underline{0.9838}} \\

\multicolumn{1}{c|}{}  &
\multicolumn{1}{l|}{{\bf MEUNet (ours)}}      & 
\multicolumn{1}{c}{\bf{28.49}/\bf{0.8484}} & 
\multicolumn{1}{c}{\bf{32.90}/\bf{0.9270}} & 
\multicolumn{1}{c}{\bf{37.64}/\bf{0.9648}} &
\multicolumn{1}{c}{\bf{42.39}/\bf{0.9842}} \\

\toprule[1pt]
\end{tabular}
}
\end{table*}

\begin{figure*}[t]
\centering
\includegraphics[width=.9\linewidth]{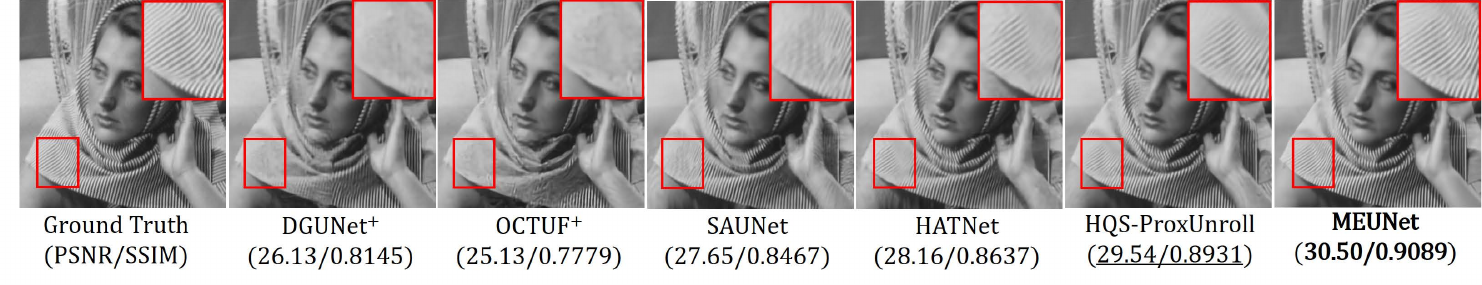}
\caption{The comparison of visualization results (\texttt{Barbara} in Set 11 ) at $SR=10\%$ of different methods. }
\label{fig: visual_simu}
\end{figure*}

Having aggregated the essential image information into $\Vmat_{YD}$, the second stage precisely propagates this guidance back to every pixel location in the original, full-resolution feature map. First, We project the original, full-resolution feature map $\Fmat_{in} \!\in\! \mathbb{R}^{N_H\times N_W\times N_C}$ to generate a new set of Queries vectors, $\Qmat_H \!\in\! \mathbb{R}^{N_HN_W \times N_d}$. Then, a backward attention map is computed by high-resolution queries and keys map from measurement.
\begin{equation}
\begin{aligned}
\label{eq: batten}
\Amat_{HY} = \textstyle {\rm softmax}(\frac{\Qmat_H\Kmat^\Tmat_Y}{\sqrt{d_k}}). 
\end{aligned}
\end{equation}
This step can be intuitively understood as each pixel location in the feature map 
is looking for the most relevant piece of information for restoration with given content.
The backward attention map is then used to weight and aggregate the summarized features from $\Vmat_{YD}$, producing the refined feature sequence $\Fmat_{r}$:
\begin{equation}
\begin{aligned}
\label{eq: bmatten}
\Fmat_r = \Amat_{HY} * \Vmat_{YD}.
\end{aligned}
\end{equation}
The resulting sequence $\Fmat_{r} \!\in\! \mathbb{R}^{N_HN_W \times N_d}$ is then reshaped back to the input feature dimensions.
This backward pass ensures that the global information contained in $\Vmat_{YD}$ is distributed precisely and adaptively across the entire spatial domain of the feature map, rather than being broadcast uniformly. It allows for a fine-grained, pixel-level modulation based on  global measurement constraints.
\section{Results and Analysis}
\label{sec:results}
Following previous works~\cite{shi2019image,shi2019scalable,song2021memory,zhang2020amp,shen2022transcs,mou2022deep,song2023optimization,wang2023saunet, qu2024dual}, 400 images from BSD500~\cite{5557884} are employed as the training dataset. Data augmentation operations, including random horizontal flipping, scaling, and cropping, are performed to generate 20,000 images as the training dataset.
All models are trained through 200 epochs with learning rate ${1\times10^{-3}}$ and then fine-tuned through 20 epochs with learning rate ${1\times10^{-4}}$ using Adam optimizer (${\beta_1} = 0.9$, ${\beta_2} = 0.999$).
all the sensing matrices in AKCS model are set to be learnable for fair comparison in simulation.
For testing on synthetic data, we evaluate the proposed method with different sampling ratios (SRs) $\{4\%,10\%,25\%,50\% \}$ on a commonly-used Set11 dataset.
Peak Signal to Noise Ratio (PSNR) and Structural Similarity (SSIM) are used to estimate the performance in our experiments. 
For real data, we build a real SPI system to verify the effectiveness of our method. The learned sensing matrices are used in real experiment.
\begin{figure*}[t]
\centering
\includegraphics[width=0.75\linewidth]{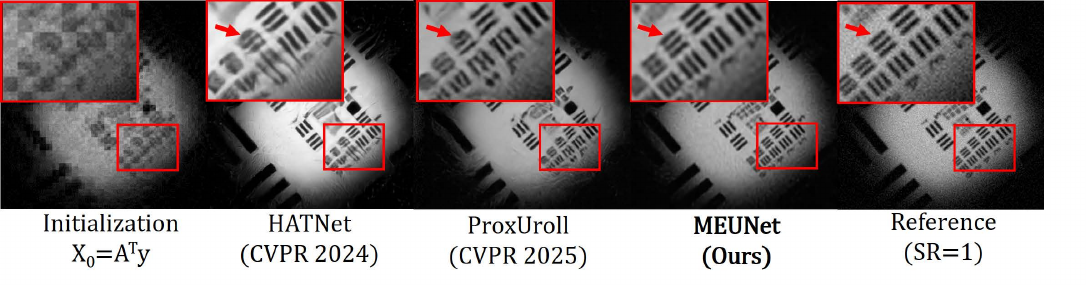}
\caption{The comparison of real data at $SR=10\%$ of different methods. }
\label{fig: visual_real}
\end{figure*}

\subsection{Results on Synthetic Data}
We make a comprehensive comparison with previous methods to evaluate the performance of proposed MEUNet, including ReconNet~\cite{kulkarni2016reconnet}, ISTA-Net$^+$~\cite{zhang2018ista}, CSNet$^+$~\cite{shi2019image}, SCSNet~\cite{shi2019scalable}, OPINENet$^+$~\cite{zhang2020optimization}, AMP-Net~\cite{zhang2020amp}, TransCS~\cite{shen2022transcs}, MADUN~\cite{song2021memory}, DGUNet$^+$~\cite{mou2022deep}, OCTUF$^+$~\cite{song2023optimization}, SAUNet~\cite{wang2023saunet}, HATNet~\cite{qu2024dual} and ProxUnroll~\cite{Wang_2025_CVPR}.
ReconNet and ISTA-Net$^+$ are two representative methods using random Gaussian sensing matrix and the other methods adopt the learnable sensing matrix.  SAUNet~\cite{wang2023saunet}, HATNet~\cite{qu2024dual} and ProxUnroll~\cite{Wang_2025_CVPR} are the most recent works that adopt KCS model.
Tab.~\ref{tab: result} presents the average PSNR/SSIM of different methods.The proposed MEUNet outperforms previous methods at all SRs. 
Fig.~(\ref{fig: overall}) presents the visualization comparison on the reconstruction results of our MEUNet and previous competitive methods. The proposed method showcases the best results and also improvement in details and textures, as highlighted in the zoom-in regions.

\subsection{Results on Real Data}

We further demonstrate the performance of the proposed method on real data. We use our SPI system to capture real data of a resolution target with $256\times256$ pixels, and then they are reconstructed by HATNet~\cite{qu2024dual}, ProxUnroll~\cite{Wang_2025_CVPR} and our proposed MEUNet, respectively.
The reconstructed results are presented in~\ref{fig: visual_real}. 
The reference images are captured by the SPI (SR=1).
Full-sampling SPI can be formulated as $\xv = \Amat^\top\yv$ s.t. $\yv = \Amat\xv$, where $\Amat \!\in\!\mathbb{R}^{N\times N}$ is an orthogonal Hadamard matrix.
In theory, full-sampled image using Hadamard matrix is lossless because of its orthogonality, thus the full-sampled image can also be regarded as a reference for comparison.
Compared with the other methods, our proposed MEUNet showcases the best reconstruction results. 

\subsection{Ablation Study}
\noindent{\bf Different Components of MEUNet.}
To quantitatively analyze the effect of different components, we perform ablation experiments on Set11 dataset at SR = $10\%$. The proposed MEUNet is mainly powered by the following designs: asymmetric Kronecker compressive sensing (AKCS) model, spatial convolution block (SCB), channe self-attention block (CSAB), measurement-aware cross-attention block (MACA).
The average PSNR and SSIM are shown in Tab.\ref{tab: ablation}. Baseline model (a) contains all the components and presents the best result of 32.90 dB/0.9270. Model (b) without AKCS, there is an
average 0.55 dB /0.0073 degradation on PSNR and SSIM. The absence of AKCS model in MEUNet leads to the most severe degradation in reconstruction, which demonstrates the importance of incoherent sensing in CS, and also the effectiveness of the proposed AKCS model.
Towards model (c) without SCB in convolution-Transformer block, there is an
average 0.27 dB/0.0031 degradation on PSNR and SSIM, which demonstrates the performance of proposed SCB in DUN.
Towards model (d) without CSAB in convolution-Transformer block, there is an
average 0.48 dB/0.0065 degradation on PSNR and SSIM.
Towards model (e) without MACA, there is an
average 0.19 dB/0.0027 degradation on PSNR and SSIM, meaning that the proposed MACA, as an plug-and-play part in DUN, is effective for CS reconstruction.

\begin{table}[H]
\centering
\caption{Ablation study for different components in MEUNet.} 
\label{tab: ablation}
\renewcommand\tabcolsep{6.5pt}
\scalebox{0.75}{
\begin{tabular}{|c|c|c|c|c|c|c|}
\hline
Model & AKCS & SCB & CSAB & MACA & PSNR (dB) & SSIM   \\ \hline
(a) & \checkmark  & \checkmark  & \checkmark & \checkmark  & {\bf 32.90} & {\bf 0.9270} \\  \hline 
(b) & &  \checkmark   &  \checkmark  &  \checkmark  & 32.35 & 0.9194 \\ \hline
(c) & \checkmark  &      &    \checkmark    & \checkmark  & 32.63 & 0.9239 \\ \hline
(d) & \checkmark  & \checkmark &  & \checkmark & 32.42 & 0.9205 \\ \hline
(e) & \checkmark  & \checkmark  & \checkmark &   & 32.71 & 0.9243 \\  \hline 
\end{tabular}
}\end{table} 

 

\noindent{\bf Comparison of computational complexity.}
Then we further make a comparison on average PSNR, the number of parameters, FLOPs, and inference time on image size of $256 \times\ 256$ of different methods, as presented in Tab.~\ref{tab: inference}. The results also demonstrate the superiority of the proposed method on performance and efficiency.

\begin{table}[th]
\centering
\caption{Comparisons on average PSNR, parameters, FLOPs, and inference time on Set11 dataset of different methods.} 
\label{tab: inference}
\renewcommand\tabcolsep{4.8pt}
\scalebox{0.75}{
\begin{tabular}{|c|c|c|c|c|c|}
\hline
Methods & CPPNet     & HATNet    &  ProxUnroll   & MEUNet    \\ 
\hline
PSNR (dB)  & 30.73  & 32.26 & \underline{32.55} & {\bf 32.90}  \\ 

\hline
FLOPs (G)   & 153.47  & 494.42   & {\bf 107.73}  & \underline{141.27} \\ 
\hline
Params (M)  & 12.31     & 31.28    & \underline{3.90}    & {\bf 2.59}  \\
\hline
InferenceTime (s) & 0.41  & 0.60 & \underline{0.27} & {\bf 0.21}  \\ 
\hline
\end{tabular}}
\end{table}

\section{Conclusion}
\label{sec:conclusion}
In this paper, we try to break the measurement barrier in both sensing and reconstruction ends. Considering the drawback introduced by KCS, we propose AKCS model in DUN, which breaks the limitation of highly structured distribution introduced by Kronecker product, thus largely leveraging the capability of expressiveness of DUN and significantly promoting the reconstruction quality with almost no increase of computational complexity. In addition, for the first time,  
we explore the implicit prior embedded in the measurement using a plug-and-play measurement-aware cross-attention module in DUN and demonstrate its effectiveness in promoting the reconstruction accuracy. The proposed AKCS model and MACA are combined with efficient Convolution-Transformer design in DUN leads to both SOTA performance and high efficiency compared to previous works. Furthermore, we also demonstrate the feasibility of the proposed method in real-world CS system, especially the AKCS model, would be a potential solution to further optimize the modulation masks for CS detection, thus promoting the application of deep-learning models in real imaging systems.

{\small
\bibliographystyle{ieeenat_fullname}
\bibliography{11_references}
}


\end{document}